\documentclass[twocolumn, 10pt]{article}
\usepackage{graphicx} 

\usepackage{cite}
\usepackage{amsmath,amssymb,amsfonts}
\usepackage{algorithmic}
\usepackage{graphicx}
\usepackage{textcomp}
\usepackage{xcolor}
\usepackage{float}
\usepackage{comment}
\usepackage{enumitem}
\usepackage{amssymb}
\usepackage{mathrsfs}
\usepackage{amsmath}
\usepackage{geometry}
\usepackage{setspace}

\usepackage[pdfborder={0 0 0}, colorlinks=true, linkcolor=black, citecolor=blue, urlcolor= black]{hyperref}
\def\BibTeX{{\rm B\kern-.05em{\sc i\kern-.025em b}\kern-.08em
    T\kern-.1667em\lower.7ex\hbox{E}\kern-.125emX}}

\makeatletter
\renewenvironment{abstract}{%
  \par\small
  \noindent\textbf{Abstract—}\ignorespaces
}{%
  \par\normalsize\bfseries
}
\makeatother

\usepackage{tikz}

\newcommand{\copyrightnotice}{
\begin{tikzpicture}[remember picture,overlay]
\node[anchor=south,yshift=10pt] at (current page.south) {
\fbox{\parbox{\dimexpr\textwidth-2\fboxsep-2\fboxrule\relax}{
\footnotesize
© 2024 IEEE. Personal use of this material is permitted. Permission from IEEE must be obtained for all other uses, in any current or future media, including reprinting/republishing this material for advertising or promotional purposes, creating new collective works, for resale or redistribution to servers or lists, or reuse of any copyrighted component of this work in other works.

This is the Author Accepted Manuscript version of a paper accepted for publication in the \href{https://doi.org/10.1109/ICSTCC62912.2024}{28th International Conference on System Theory, Control and Computing (ICSTCC)}, 2024. The published version is available via DOI: \href{https://doi.org/10.1109/ICSTCC62912.2024.10744717}{10.1109/ICSTCC62912.2024.10744717}.
}}
};
\end{tikzpicture}
}

\usepackage{titlesec}

\renewcommand{\thesection}{\Roman{section}}
\renewcommand{\thesubsection}{\Alph{subsection}}
\renewcommand{\thesubsubsection}{\arabic{subsubsection}}

\titleformat{\section}
  {\normalfont\filcenter\MakeUppercase}
  {\thesection.}
  {0.6em}
  {}

\titleformat{\subsection}
  {\normalfont\itshape}
  {\thesubsection.}
  {0.6em}
  {}

\titleformat{\subsubsection}[runin]
  {\normalfont\itshape}
  {\thesubsubsection)}
  {0.6em}
  {}
  [.\ ] 

\titlespacing*{\section}{0pt}{1.2ex plus 0.3ex minus 0.2ex}{0.8ex}
\titlespacing*{\subsection}{0pt}{1.0ex plus 0.3ex minus 0.2ex}{0.6ex}
\titlespacing*{\subsubsection}{0pt}{0.8ex plus 0.2ex minus 0.2ex}{0.6em}

\begin{document}

\makeatletter
\renewcommand{\maketitle}{%
  \twocolumn[{%
    \begin{@twocolumnfalse}
      \vspace*{-1.2em}
      \begin{center}
        {\huge\bfseries \@title \par} 
        \vspace{0.8em}
        {\normalsize
        \begin{tabular}{@{}ccc@{}}
          \begin{tabular}{@{}c@{}}
            \textbf{Zeyad Gamal}\\
            \textit{Mechatronics Engineering Department}\\
            \textit{The German University in Cairo}\\
            Cairo, Egypt\\
            \texttt{zeyad.abdrabo@student.guc.edu.eg}
          \end{tabular}
          &
          \begin{tabular}{@{}c@{}}
            \textbf{Youssef Mahran}\\
            \textit{Mechatronics Engineering Department}\\
            \textit{The German University in Cairo}\\
            Cairo, Egypt\\
            \texttt{youssef.mahran@student.guc.edu.eg}
          \end{tabular}
          &
          \begin{tabular}{@{}c@{}}
            \textbf{Ayman El-Badawy}\\
            \textit{Mechatronics Engineering Department}\\
            \textit{The German University in Cairo}\\
            Cairo, Egypt\\
            \texttt{ayman.elbadawy@guc.edu.eg}
          \end{tabular}
        \end{tabular}\par
        }
      \end{center}
      \vspace{0.8em}
    \end{@twocolumnfalse}
  }]%
}
\makeatother

\title{Control of a Twin Rotor using Twin Delayed Deep Deterministic Policy Gradient (TD3)
}

\date{}

\maketitle
\copyrightnotice
\begin{abstract}
\bfseries This paper proposes a reinforcement learning (RL) framework for controlling and stabilizing the Twin Rotor Aerodynamic System (TRAS) at specific pitch and azimuth angles and tracking a given trajectory. The complex dynamics and non-linear characteristics of the TRAS make it challenging to control using traditional control algorithms. However, recent developments in RL have attracted interest due to their potential applications in the control of multirotors. The Twin Delayed Deep Deterministic Policy Gradient (TD3) algorithm was used in this paper to train the RL agent. This algorithm is used for environments with continuous state and action spaces, similar to the TRAS, as it does not require a model of the system. The simulation results illustrated the effectiveness of the RL control method.  Next, external disturbances in the form of wind disturbances were used to test the controller's effectiveness compared to conventional PID controllers. Lastly, experiments on a laboratory setup were carried out to confirm the controller's effectiveness in real-world applications.
\end{abstract}

\section{Introduction}

The Twin Rotor Aero-dynamic System (TRAS) is a laboratory setup designed for control experiments. It is a non-linear cross-coupled system that resembles the dynamics of a helicopter as it features two perpendicular rotors that allow rotation in the vertical and horizontal planes. Designing a controller for the TRAS to stabilize at different positions and to track trajectories whilst being stable is a difficult task due to the complexity of its mathematical model.

In the literature, several studies have been carried out to test different types of controllers for the TRAS. Fuzzy logic control was used, and simulations in addition to hardware experiments showed that the system was able to reach the desired positions efficiently \cite{juang2011hybrid}. An H$_\infty$ controller was proposed to stabilize the TRAS \cite{ahmed2009robust}. This was done through decoupling the system dynamics using Hadamard Weights. The simulation results proved the effectiveness of the controller. Hardware
implementation was carried out and yielded similar results to the simulations. Another robust H$_\infty$ controller was implemented to track the reference signals for the pitch and azimuth angles while taking into consideration input disturbances, measurement noise, and ten parametric uncertainties in the mathematical model \cite{hassan2020robust}. Both simulations and physical experiments verified the performance of the controller. Sliding mode control was used to provide robustness to external disturbances and unmodelled dynamics of the system \cite{rashad2017sliding}. Several simulations were done and compared with other control algorithms to show that this approach, provides less tracking error with lower control effort. A Robust and optimal Model Predictive Controller (MPC) was used to linearize and stabilize the system \cite{ulasyar2015robust}. The simulations gave better results than the traditional Linear Quadratic Regulator (LQR) approach.

However, in recent years, there have been huge developments in the fields of artificial intelligence and machine learning which has led to the utilization of reinforcement learning techniques in control applications instead of traditional control methods. Several studies have been presented to show how reinforcement learning is used to control multirotors particularly quadrotors to fly autonomously. 

A path-following and stabilization agents for the quadrotor was proposed by training the agents to map the environment states into motor commands \cite{shehab2021low}. The paper utilized a full-state representation of the quadrotor model, including position, orientation, and linear and angular velocities. The agent was trained using the Twin Delayed Deep Deterministic Policy Gradient (TD3) algorithm which is a model-free, deterministic, and off-policy actor-critic method that utilizes neural networks. The simulations proved the effectiveness of the stabilization agent and the successful trajectory tracking for the path-following agent. An actor-critic technique was also proposed using two neural networks \cite{hwangbo2017control}. Simulations and real-world experiments were carried out to verify the technique. Two RL agents were trained using the TD3 algorithm to address the quadrotor-related path following (PF) and obstacle avoidance problems (OA) \cite{mokhtar2023autonomous}. The PF agent's goal is to map observations into motor commands to follow a predetermined path. The OA agent's goal is to modify the tracking error information to ensure that the followed path is free from any obstacles before sending the information to the PF agent. This control framework was tested by following different paths and avoiding multiple obstacles using simulations. 

The objective of this paper is to design a controller using the TD3 algorithm to stabilize the TRAS at a given reference position and to track a trajectory with high accuracy. External disturbance added to the input signal in the form of wind disturbance is also used to verify the robustness of the agent compared to conventional Proportional-Integral-Derivative (PID) gain controllers. Practical experiments were also carried out to show the successful implementation of the controller. This control framework has not been proposed in the literature so far for the TRAS.

\section{Methodology}
\subsection{Simulation Environment}
The software used for simulation is MATLAB/Simulink. The Simulink model of the TRAS is obtained from \cite{inteco}. The model serves as the environment the RL agent interacts with. The environment takes as an input two control signals from the agent as shown in Fig. \ref{controller} representing input voltages for each of the vertical and horizontal rotors respectively. 
\begin{equation}
      I = [U_{v},U_{h}]
    \label{eq1}
\end{equation}
The environment then outputs pitch angle, pitch angular velocity, azimuth angle, and azimuth angular velocity respectively.
\begin{equation}
      O = [\theta, w_{\theta}, \psi, w_{\psi}]
    \label{eq2}
\end{equation}
The environment outputs in Eq. \ref{eq2} are fed back to the RL agent as an input to the observation space as shown in Fig. \ref{controller}.

\subsection{Control Algorithm}
\begin{figure}[!h]
\centerline{\includegraphics[width=0.475\textwidth]{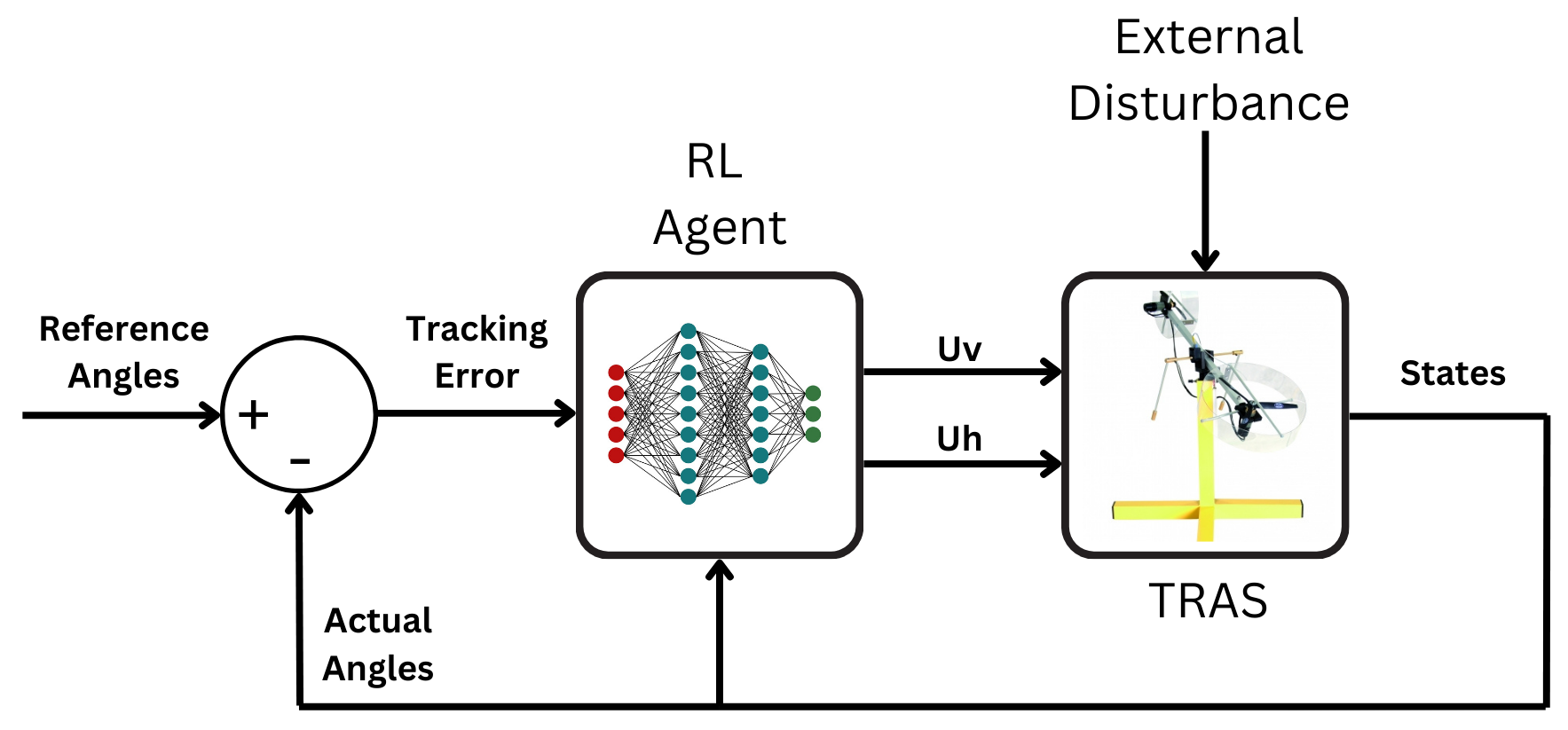}}
\caption{Block diagram of control algorithm}
\label{controller}
\end{figure}
\begin{figure}[H]
\centerline{\includegraphics[width=0.475\textwidth]{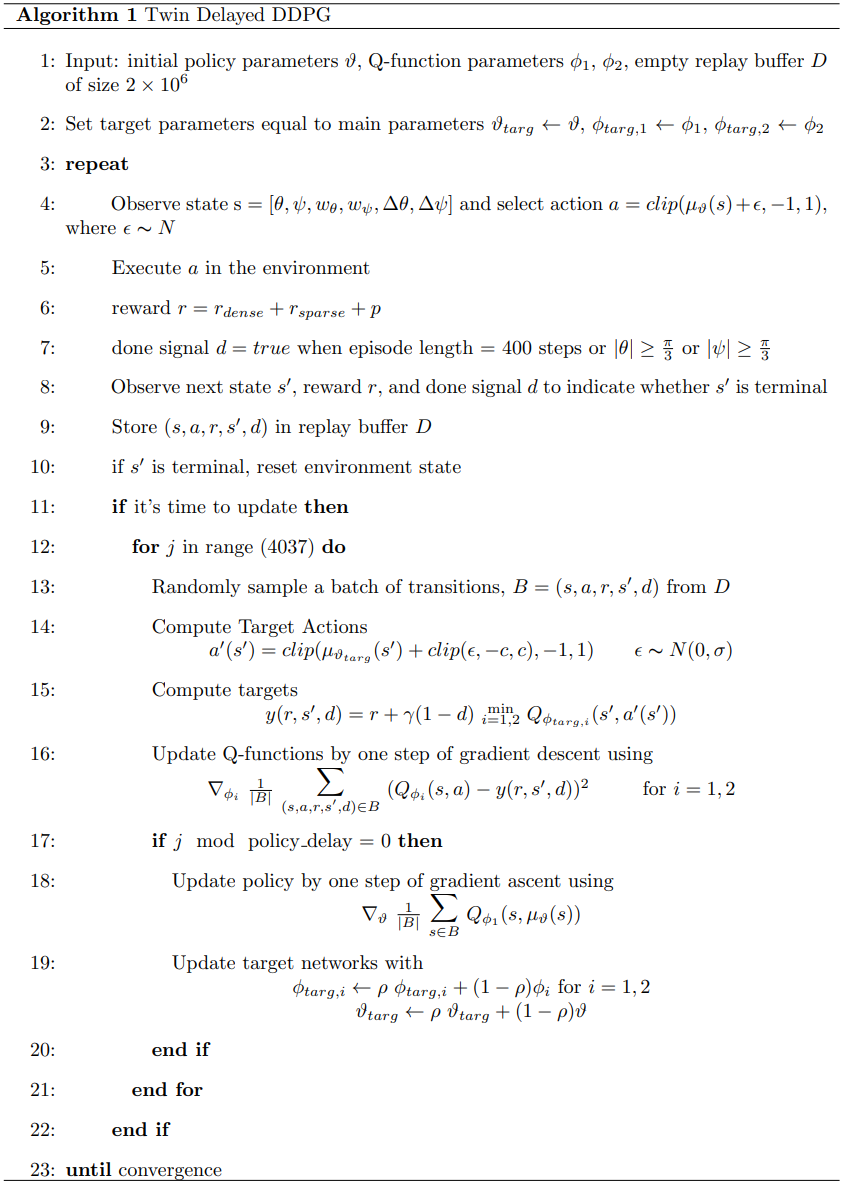}}
\caption{TD3 algorithm pseudocode \cite{fujimoto2018addressing}}
\label{TD3}
\end{figure}

A deep reinforcement learning-based controller has been developed in this paper as shown in Fig. \ref{controller}. The RL agent receives as inputs the tracking error, the difference between the desired and current azimuth and pitch angles, alongside the states of the TRAS. The RL agent then sends two control signals which represent the voltages of the two rotors to the TRAS, to move in the horizontal plane and vertical plane to reach the desired reference angles.
The algorithm used to train the RL agent is the Twin Delayed Deep Deterministic Policy Gradient (TD3), a model-free and off-policy Actor-Critic technique used specifically for environments with continuous state and action spaces. The Actor-Critic technique estimates the policy by computing actions according to the current state of the environment through the actor. The critic calculates the value function evaluating the actions performed by the actor. The critic also calculates the Temporal Difference error (TD) used by the actor and critic in the training. Both actor and critic use Artificial Neural Networks (ANNs) as function approximators to allow them to learn how to map their inputs into the appropriate outputs \cite{mokhtar2023autonomous}. The architecture of the networks is presented in the following sections. A pseudocode for the TD3 algorithm used is presented in Fig. \ref{TD3}.

\subsection{Network Structure}
TD3 algorithm consists of two actor networks and four critic networks. The actor network takes as an input the state space and outputs the action while the critic network takes as an input the state space in addition to the action and outputs the Q-value. The action space consists of the two control signals of the TRAS Eq. \ref{eq1}. The state or observation space Eq. \ref{eq3} consists of 6 elements, the output vector of the TRAS model Eq. \ref{eq2} in addition to the tracking error in the form of the difference between current and reference angles.
\begin{equation}
     S = [\theta,\psi,w_{\theta},w_{\psi}, \Delta\theta,\Delta\psi]
    \label{eq3}
\end{equation}
The six networks consist of 4 layers. An input layer of 6 and 8 nodes for the actor and critic networks respectively. Two hidden layers of 400 and 300 nodes respectively with \textit{LeakyReLU} activation function and an output layer with \textit{tanh} activation function. Due to the \textit{tanh} activation in the output layer, the actor output is scaled to the range of -1 and 1. The output is then mapped to the range of -24 and 24 representing the minimum and maximum possible voltages for the motors.

\subsection{Reward Function}
Two types of reward functions are commonly used in reinforcement learning: sparse and dense reward functions. The rewards provided by sparse reward functions are usually high and available only in specific states. On the other hand, smaller rewards are given more frequently in dense reward systems. In this paper, a combination of dense and sparse rewards is used. The dense rewards are used due to the continuous nature of the TRAS system while sparse rewards are used to encourage the agent to eliminate the error when stabilizing around the reference angle.
\begin{equation}
    r_{dense} = -c*(\Delta \theta^2 + \Delta \psi^2)
    \label{eq4}
\end{equation}
 Eq. \ref{eq4} shows the dense reward function governing the agent's behavior. The dense reward function penalizes deviations away from the reference signal. ($\Delta\theta$) is the error between the current and desired pitch angle, ($\Delta\psi$) is the error between the current and desired azimuth angle, and ($c$) is a weighting term with a value of 10 obtained through trial and error. The weight of the negative dense reward function decreases as the deviation decreases.
\begin{equation}
r_{sparse} = 
    \begin{cases}
    1, & \text{if } \Delta \theta \leq 0.01 \text{ and } \Delta \psi \leq 0.01 \\
    0, & \text{otherwise}
\end{cases}
\label{sparse}
\end{equation}
The sparse reward function Eq. \ref{sparse} is only given at the state where the deviation error falls under 0.01 in both the azimuth and pitch angles. The positive nature of the sparse reward function encourages the agent to minimize the deviation. 
\begin{equation}
p = 
    \begin{cases}
    -5000, & \text{if }  \theta \geq \frac{\pi}{3} \text{ or }  \psi \geq \frac{\pi}{3} \\
    0, & \text{otherwise}
\end{cases}
\label{penalty}
\end{equation}
The sparse penalty Eq. \ref{penalty} is only given to the agent when either angle reaches its boundary. The value of p was chosen after several trials to improve the agent's performance. At first, a low value of 1000 caused the agent to terminate the episode early to avoid getting more penalties. On the other hand, a high value of 10000 caused the agent to prioritize avoiding boundaries over exploring, so it failed to track the references. The overall reward function Eq. \ref{reward} used when training the RL agent is a combination of the three reward functions.
\begin{equation}
    r = r_{dense} + r_{sparse} + p
    \label{reward}
\end{equation}

\subsection{TD3 Hyperparameters}
    \begin{table}[h]
     \caption{TD3 Hyperparameters used in training}
        \centering
        \begin{tabular}{lll}
            \hline
            \textbf{Parameter} & \textbf{Symbol} & \textbf{Value} \\
            \hline
            Sampling Time & $t_s$ & 0.05 seconds\\
            Maximum steps per episode & - & 400\\
            Critic Learning Rate & $\alpha$ &$1 \times 10^{-4}$ \\
            Actor Learning Rate & $\alpha$ & $1 \times 10^{-4}$ \\
            Discount Factor & $\gamma$ &0.995\\
            Buffer Size & $D$ & $2 \times 10^6$ \\
            Batch Size & K & 512 \\
            Noise & $\sigma$ & 0.1\\
            Noise Decay Rate & - & 1 $\times 10^{-3}$\\
            Target Smooth Factor & $\tau$ & 0.005\\
            Target Update Frequency & - & 10\\
            \hline
            \label{table1}
        \end{tabular}
        \label{TD3 Hyperparameters}
    \end{table}
The hyperparameters of TD3 have a strong effect on its effectiveness. Small changes in these parameters can have a big impact on performance results. The sampling time was set to a low value of 0.05 seconds so the agent could evaluate the states more often and make frequent changes to its policy to adapt quickly to the changes in the states. The learning rates were set to a low value to ensure stability in the training and that it will eventually converge even though training time will be longer. The discount factor encourages the agent to favor future rewards and not just immediate rewards when calculating its action. A noise ($\sigma$) is added to encourage the agent to explore different approaches in the learning process. The hyperparameters used was obtained through trial and error. Table. \ref{table1} shows the hyperparameters used in training.

\subsection{Simulation Objective and Conditions}
The proposed agent was trained twice with the same hyperparameters and reward functions. The initial training aimed to start from the initial position and stabilize at a certain preset position. The initial ($\psi$) and ($\theta$) were set to (0,0). A reset function was implemented to change the reference pitch angle and azimuth angle at the beginning of each episode. This was done by choosing a random element from an array that contains the values [0.4,0.2,-0.4,-0.2] for the azimuth and another array with values [0.25,0,-0.25,-0.3] for the pitch and setting the randomly chosen values as the references. This is repeated at the start of each episode during training so the agent will be able to track any references.

The obtained agent underwent further training. The second training aimed to track a certain trajectory. The agent starts from the initial position of (0,0) and proceeds to follow the trajectory. 

Due to the continuous nature of the environment and to reduce the training time, the number of states the agent needed to explore was reduced by limiting the state space. This was accomplished by setting boundary conditions that would terminate the episode as follows:
\begin{itemize}
    \item $|\theta| \geq \frac{\pi}{3}$
    \item $|\psi| \geq \frac{\pi}{3}$
\end{itemize}
The episode terminates when either the azimuth or pitch exceeds $60^\circ$. This is done as the experimental model is also limited to the same value to avoid crashing.

\subsection{Wind Disturbance Model}
To simulate the effects of the wind external disturbances, the Dryden Wind Turbulence Model is used to add external disturbances to the system to be able to test the robustness of the controller in handling changes. The external disturbances are added to the environment itself as shown in Fig. \ref{controller}. The controller then tries to eliminate the effect of the disturbances and compensate for it. The Dryden Wind Turbulence Model is widely used in the applications of aerospace as it provides a realistic representation of the turbulence affecting aircraft \cite{moorhouse1982background}. In this model, the wind velocity waveform is generated by passing band-limited white Gaussian noise with unity standard deviation through the filters shown in Eqs. \ref{filter1}, \ref{filter2} and \ref{filter3}.
\begin{equation}
G_u(s) = \frac{\hat{v}_{wx}(s)}{\eta_u(s)} = \sigma_u \sqrt{\frac{2L_u}{\pi V}} \frac{1}{1 + \frac{L_u}{V}s}
\label{filter1}
\end{equation}

\begin{equation}
G_v(s) = \frac{\hat{v}_{wy}(s)}{\eta_v(s)} = \sigma_v \sqrt{\frac{L_v}{\pi V}} \frac{1 + \frac{\sqrt{3}L_w}{V}s}{(1+\frac{L_v}{V}s)^2}
\label{filter2}
\end{equation}

\begin{equation}
G_w(s) = \frac{\hat{v}_{wz}(s)}{\eta_w(s)} = \sigma_w \sqrt{\frac{L_w}{\pi V}} \frac{1 + \frac{\sqrt{3}L_w}{V}s}{(1+\frac{L_w}{V}s)^2}
\label{filter3}
\end{equation}
$\hat{v}_{wx}$, $\hat{v}_{wy}$ and $\hat{v}_{wz}$ are the components of the wind turbulence velocity, $\eta_u$, $\eta_v$, $\eta_3$ denote white Gaussian noise with zero mean, $\sigma_u$, $\sigma_v$, $\sigma_w$ represent turbulence intensities which depend on the wind velocity at 6 meters altitude, $L_u$, $L_v$ and $L_w$ are the turbulence scale lengths which are a function of the altitude and V is the mean air speed. The drag force acting on the body is then calculated to be transformed into torques that are resolved onto the azimuth and pitch angles.

\subsection{Network Training}
The neural networks were trained using the mentioned conditions and parameters. The training was done with the use of NVIDIA CUDA on NVIDIA RTX 2070 8G GPU, along with an Intel Core i7 processor and 32 Gigabytes of Random Access Memory (RAM). The first training was run for a total of 4037 episodes while the second training was run for 1909 episodes.

\section{Results}

In this section, the results of both the stabilization and trajectory tracking training are presented and discussed. Moreover, the response of the agent to external disturbances is presented and compared to the performance of a conventional PID controller. Lastly, the practical implementation of the agent in real life is presented.

\subsection{Stabilization Training Results}
\begin{figure}[H]
\centerline{\includegraphics[width=0.5\textwidth]{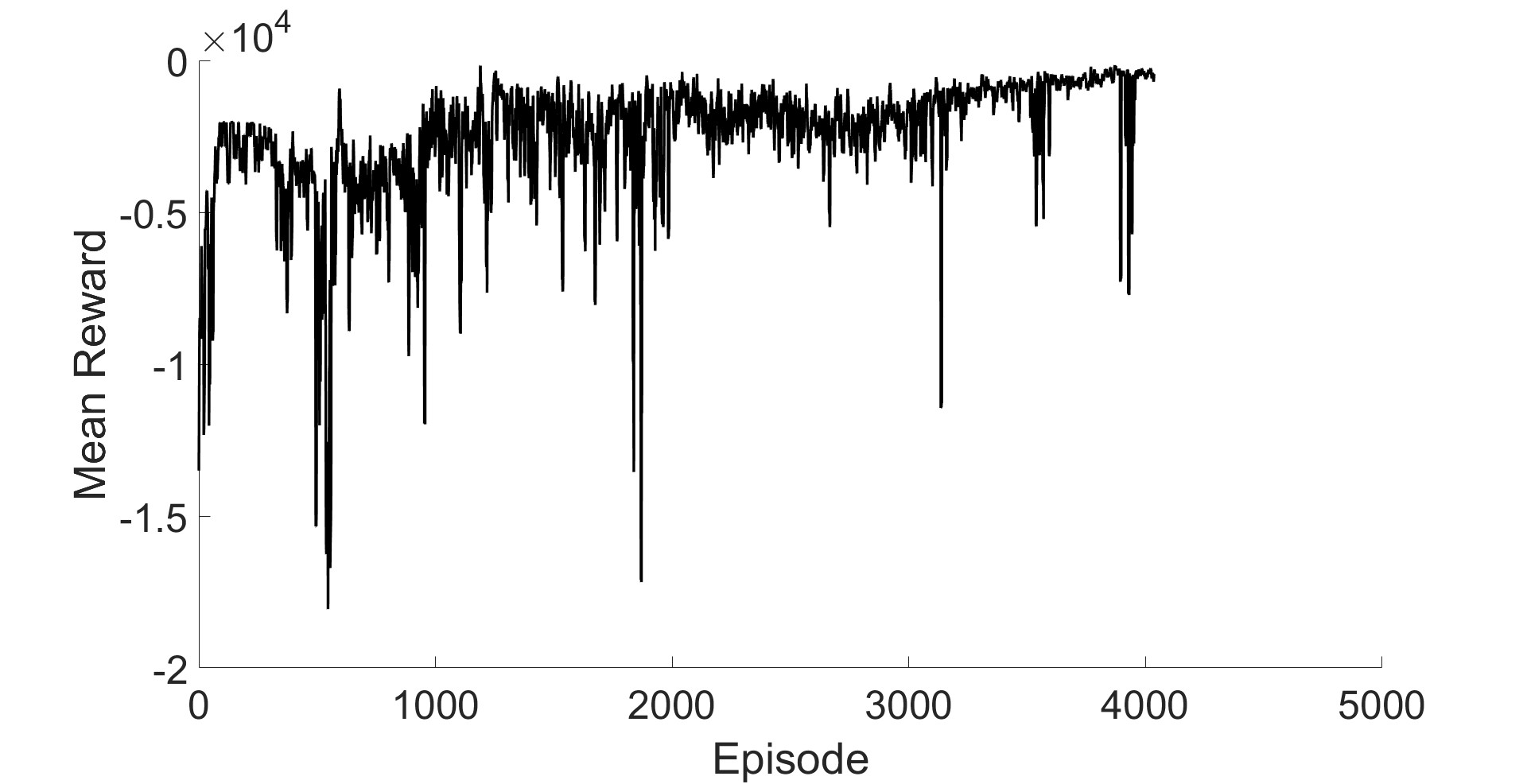}}
\caption{Mean reward per episode during stabilization training}
\label{mean}
\end{figure}

The objective of the stabilization training was to stabilize the TRAS at any given random reference position. This position is given as the pitch and azimuth angle. The TRAS should reach the given position as quickly as possible whilst ensuring a low overshoot. Fig. \ref{mean} shows the mean reward obtained per episode. Initially, the agent received large penalties as the deviations from the reference were high and the episode terminated early causing the agent to receive an even larger penalty. After almost 3000 episodes, the agent learned to minimize the deviations and avoid the termination penalty. Eventually, it managed to stabilize at the reference and receive sparse positive rewards. The mean reward increased with time which shows positive learning behavior achieved by the agent.

\begin{figure}[H]
\centerline{\includegraphics[width=0.5\textwidth]{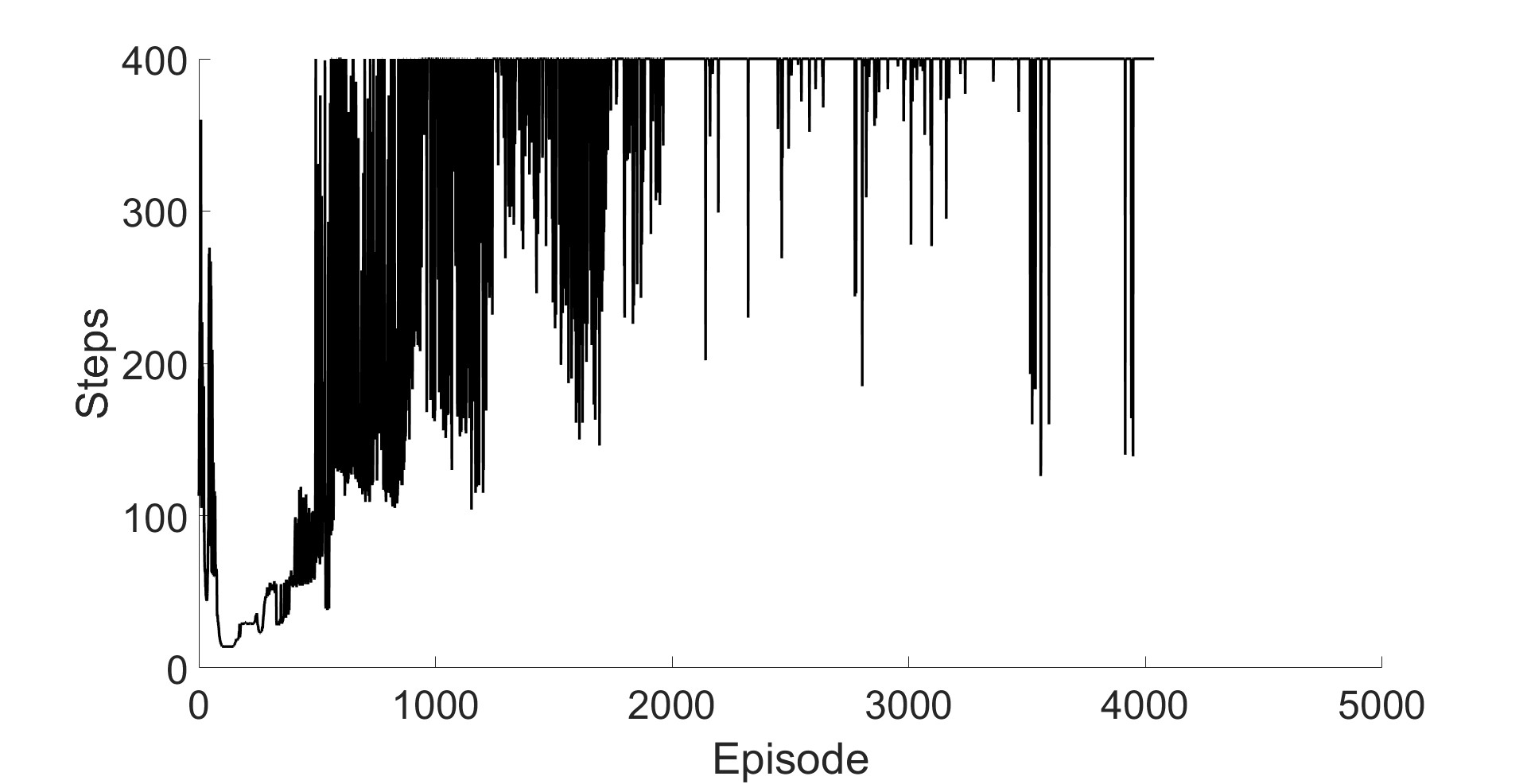}}
\caption{Steps per episode during stabilization training}
\label{steps}
\end{figure}

Fig. \ref{steps} shows the number of steps achieved by the agent in each episode. The number of steps increased with time as the agent was learning to stabilize at its reference. The number of steps is capped at 400 as the maximum number of steps per episode is set to 400 as in Table. \ref{TD3 Hyperparameters}.

\begin{figure}[H]
\centerline{\includegraphics[width=0.5\textwidth]{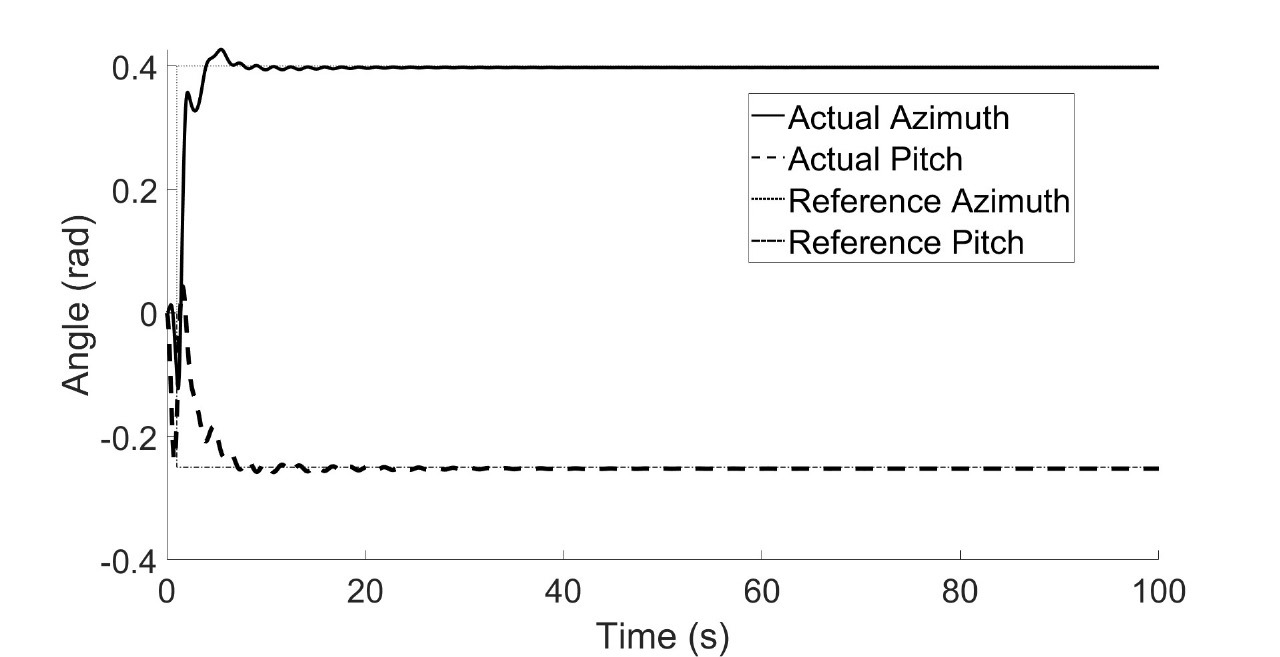}}
\caption{Pitch and azimuth response at a reference of -0.25 and 0.4 radians}
\label{pos1}
\end{figure}

Fig. \ref{pos1} shows the position of the TRAS after giving a reference azimuth angle of 0.4 radians and a pitch angle of -0.25 radians. The plots show that the TRAS reached and stabilized at the given position with very low and negligible error that is less than 0.01 radians. The overshoot was less than 0.1 radians and the settling time was about 10 seconds. 

\begin{figure}[H]
\centerline{\includegraphics[width=0.5\textwidth]{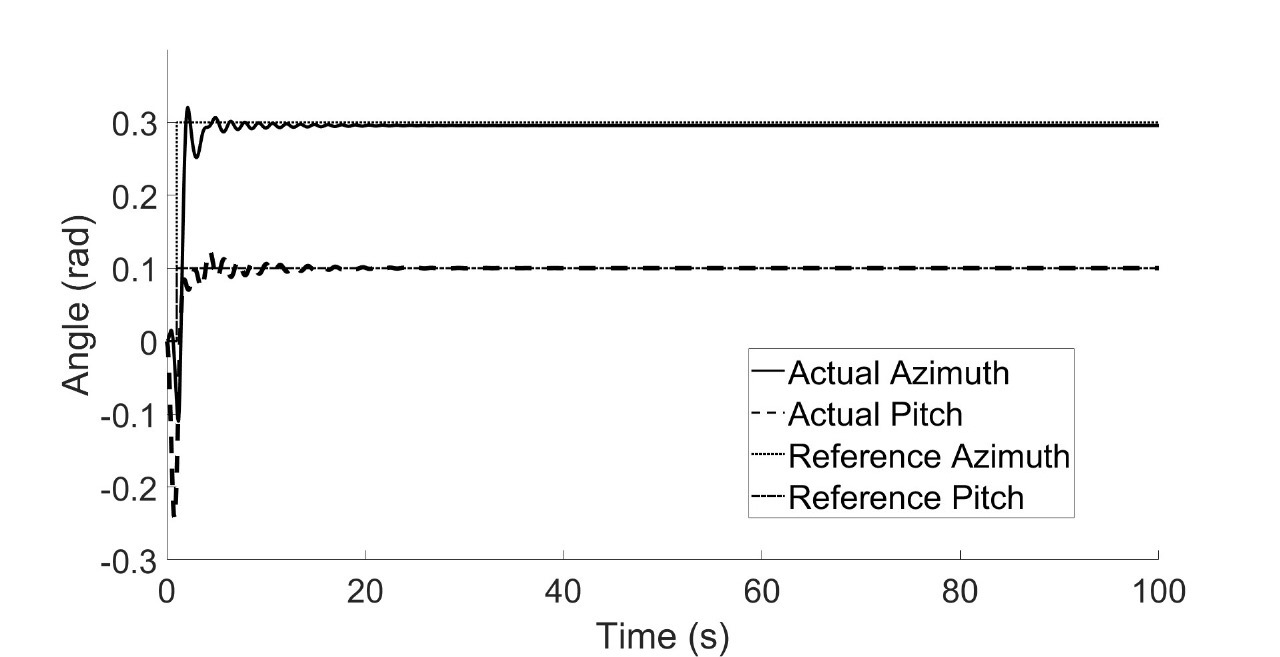}}
\caption{Pitch and azimuth response at a reference of 0.1 and 0.3 radians}
\label{pos2}
\end{figure}

Fig. \ref{pos2} shows the results obtained with another reference position. To evaluate the robustness of the agent, the agent was never trained on this reference position. The overshoot was also less than 0.01 radians and a similar settling time to the one in Fig. \ref{pos1}. This proves the agent's ability to track any reference angle with the same performance without explicitly training the agent with this angle.

\subsection{Trajectory Tracking Results}
The objective of the trajectory training was for the agent to track a certain given trajectory. The pre-trained agent from the stabilization training was further trained to track a trajectory for 1909 episodes with each episode lasting for 400 steps. During this training, both the azimuth and pitch angles were controlled simultaneously, as was done in the stabilization training.

\begin{figure}[H]
\centerline{\includegraphics[width=0.5\textwidth]{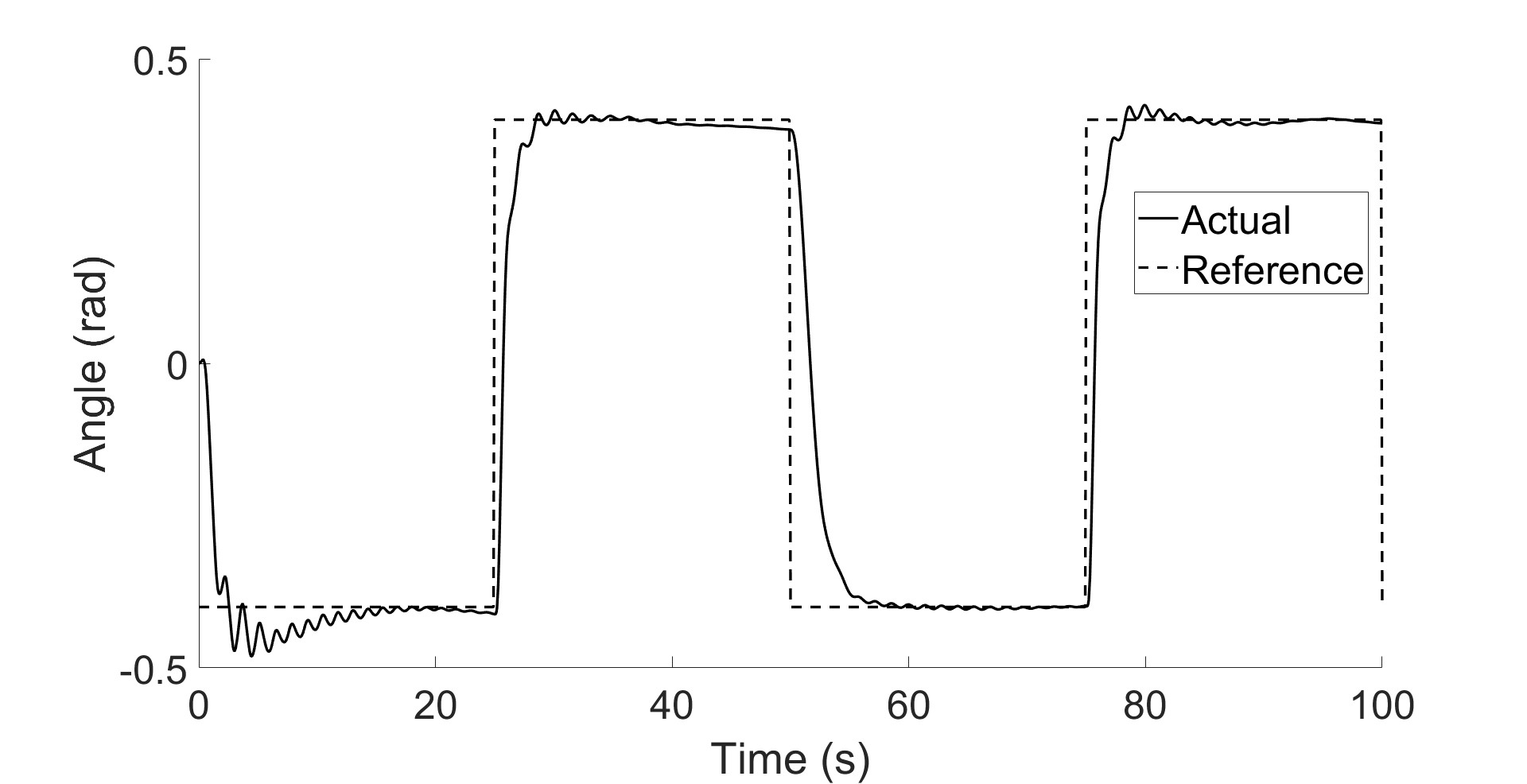}}
\caption{Azimuth tracking response for a square wave trajectory}
\label{Azimuth(Trajectory1)}
\end{figure}

The reference trajectory for the azimuth signal was chosen to be a square wave with an amplitude of 0.4 radians and frequency of $0.02$ Hz while the pitch reference trajectory was a sine wave with an amplitude of 0.25 and frequency of $0.016$ Hz. In Fig. \ref{Azimuth(Trajectory1)} the agent managed to track the given azimuth reference trajectory with a minimal steady-state error. However, as the square wave changes amplitude the agent needs some settling time with minimal overshoot.

\begin{figure}[H]
\centerline{\includegraphics[width=0.5\textwidth]{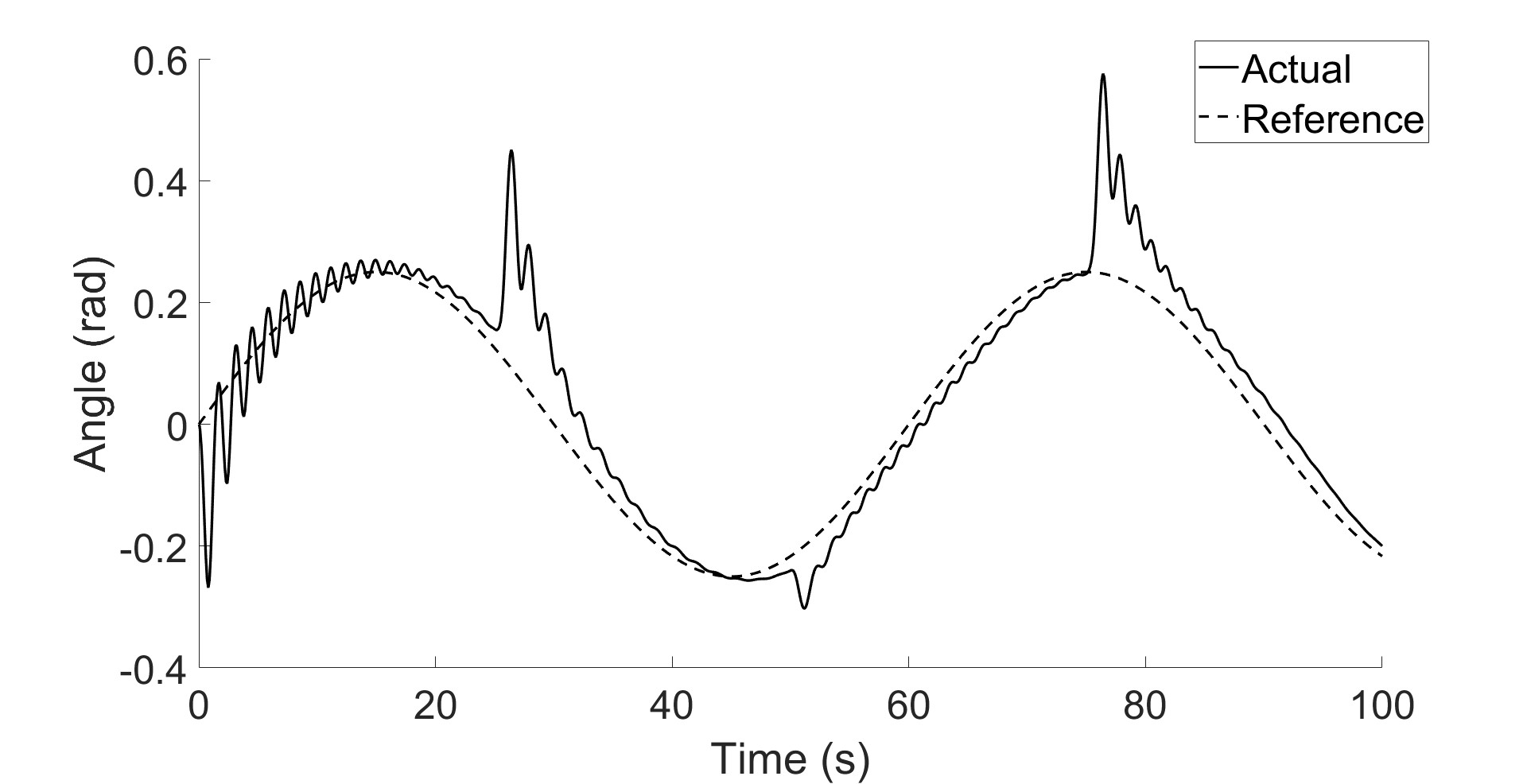}}
\caption{Pitch tracking response for a sine wave trajectory}
\label{Pitch(Trajectory1)}
\end{figure}

Fig. \ref{Pitch(Trajectory1)} shows the tracking behavior of the pitch reference trajectory. Small transient spikes in the pitch response are present. This is due to the sudden change in the reference angle of the azimuth as the system is highly coupled. Low steady-state error is present in the response of the agent for the pitch trajectory.

\begin{figure}[H]
\centerline{\includegraphics[width=0.5\textwidth]{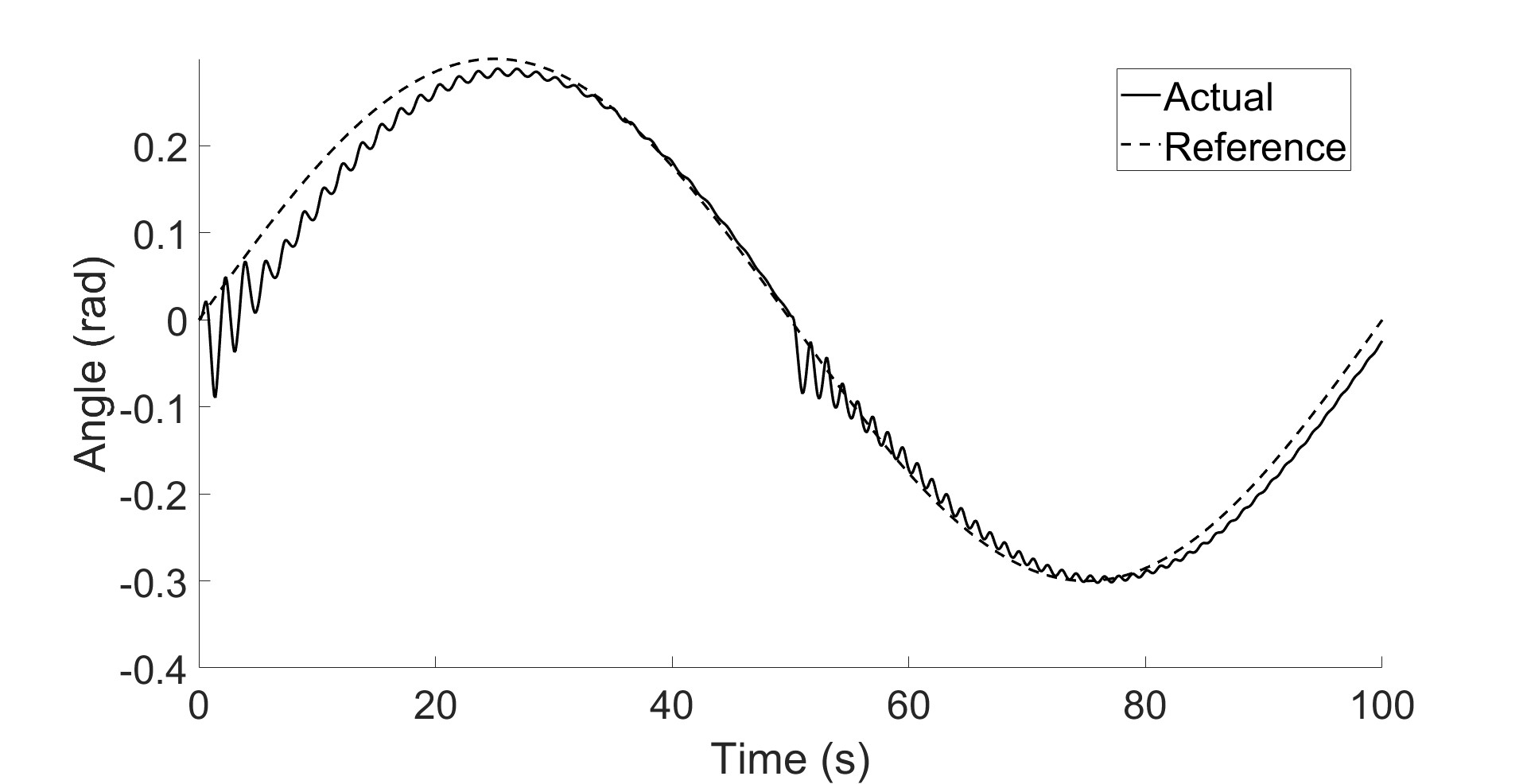}}
\caption{Azimuth tracking response for a sine wave trajectory}
\label{Azimuth(Trajectory2)}
\end{figure}

A smoother azimuth trajectory was passed to the agent in the shape of a sine wave and the square wave was set for the pitch angle to test the system coupling. Fig. \ref{Azimuth(Trajectory2)} shows successful trajectory tracking for the azimuth with minimal steady-state error. Small ripples are found in the response due to the sudden change in the pitch reference angle. This further proves the high coupling present in the system.

\begin{figure}[H]
\centerline{\includegraphics[width=0.5\textwidth]{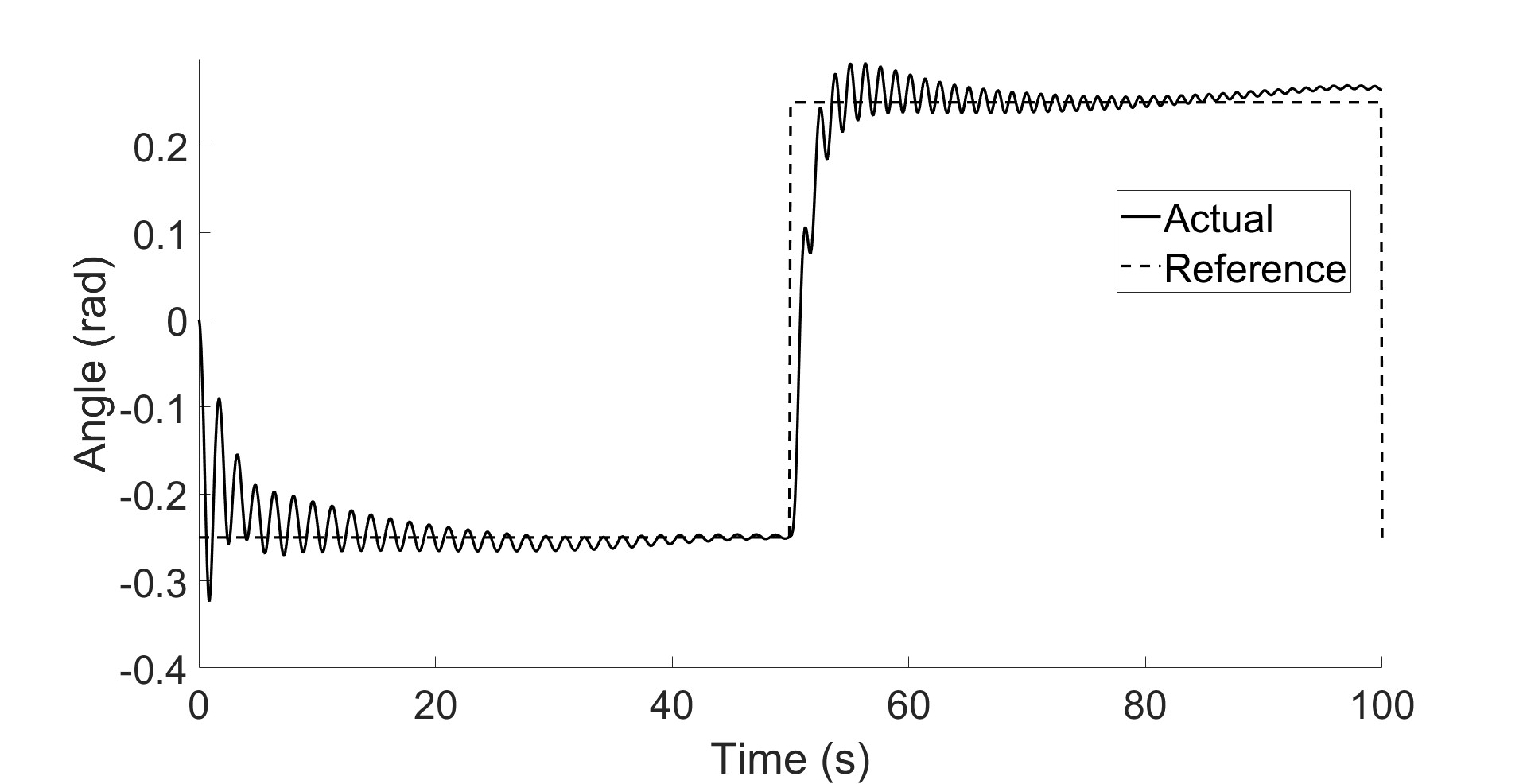}}
\caption{Pitch tracking response for a square wave trajectory}
\label{Pitch(Trajectory2)}
\end{figure}

The error in the pitch trajectory is minimal and the transient spikes are absent as shown in Fig. \ref{Pitch(Trajectory2)} due to the smoothness and the absence of sudden changes in the azimuth trajectory. A small steady-state error is present and settling time is needed when the square wave changes amplitude.

The effect of the high coupling present in the system has a larger effect on the pitch angle. This was proved through Fig. \ref{Pitch(Trajectory1)} and Fig. \ref{Azimuth(Trajectory2)}. The coupling effect causes a spike in the pitch response before settling. However, in the azimuth response, the coupling effect causes oscillations before settling again. The transient spikes in the pitch response have a larger magnitude when compared to the oscillations in the azimuth response.

\subsection{Wind Disturbance}
The next simulations were performed to test the robustness of the system against external disturbances. These disturbances are considered to be wind disturbances simulated by the Dryden Wind Turbulence Model previously discussed. The results of the RL agent are compared to the simulation results for a PID controller. The gains of the PID controller were tuned using the Ziegler-Nichols method. The gains chosen are shown in table \ref{gains}.

\begin{figure}[H]
\centerline{\includegraphics[width=0.5\textwidth]{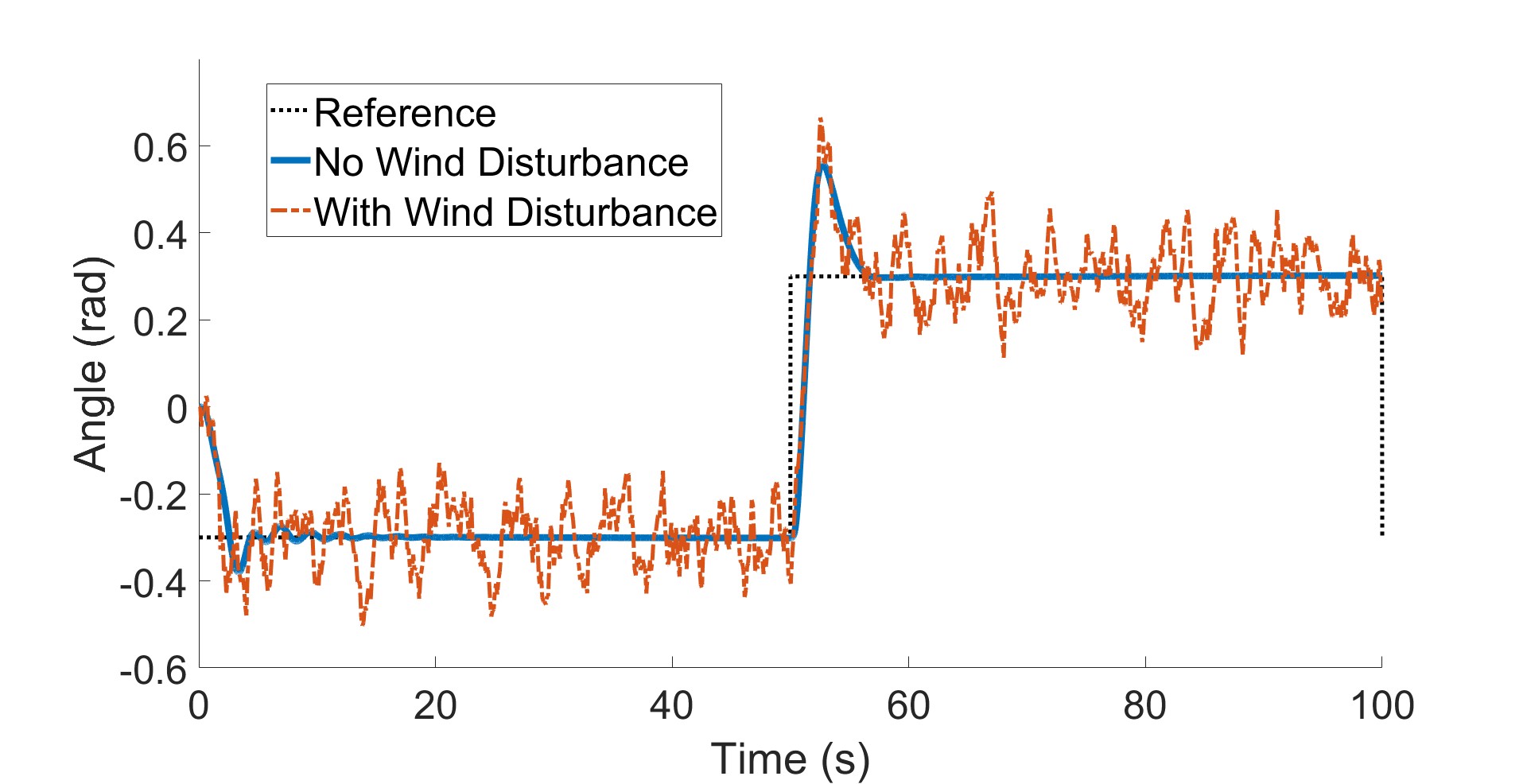}}
\caption{Azimuth trajectory tracking response using PID controller}
\label{Azimuth(pid)}
\end{figure}

\begin{table}[h]
 \caption{PID gains used in simulation}
    \centering
    \begin{tabular}{lll}
        \textbf{Gains} & \textbf{Azimuth} & \textbf{Pitch} \\
        \hline
        $K_P$ & 0.8 & 0.42\\
        $K_I$ & 0.35 & 0.684\\
        $K_D$ & 0.715 & 0.365 \\
        \hline
        
    \end{tabular}
    \label{gains}
\end{table}
Fig. \ref{Azimuth(pid)} shows the azimuth response when tracking a trajectory using the PID controller with and without external wind disturbances. Without disturbances, the system was able to track the reference signals with almost no steady-state error however, with disturbances, the PID controller failed to track the trajectory and stabilize in time as there were many oscillations in the response.

\begin{figure}[H]
\centerline{\includegraphics[width=0.5\textwidth]{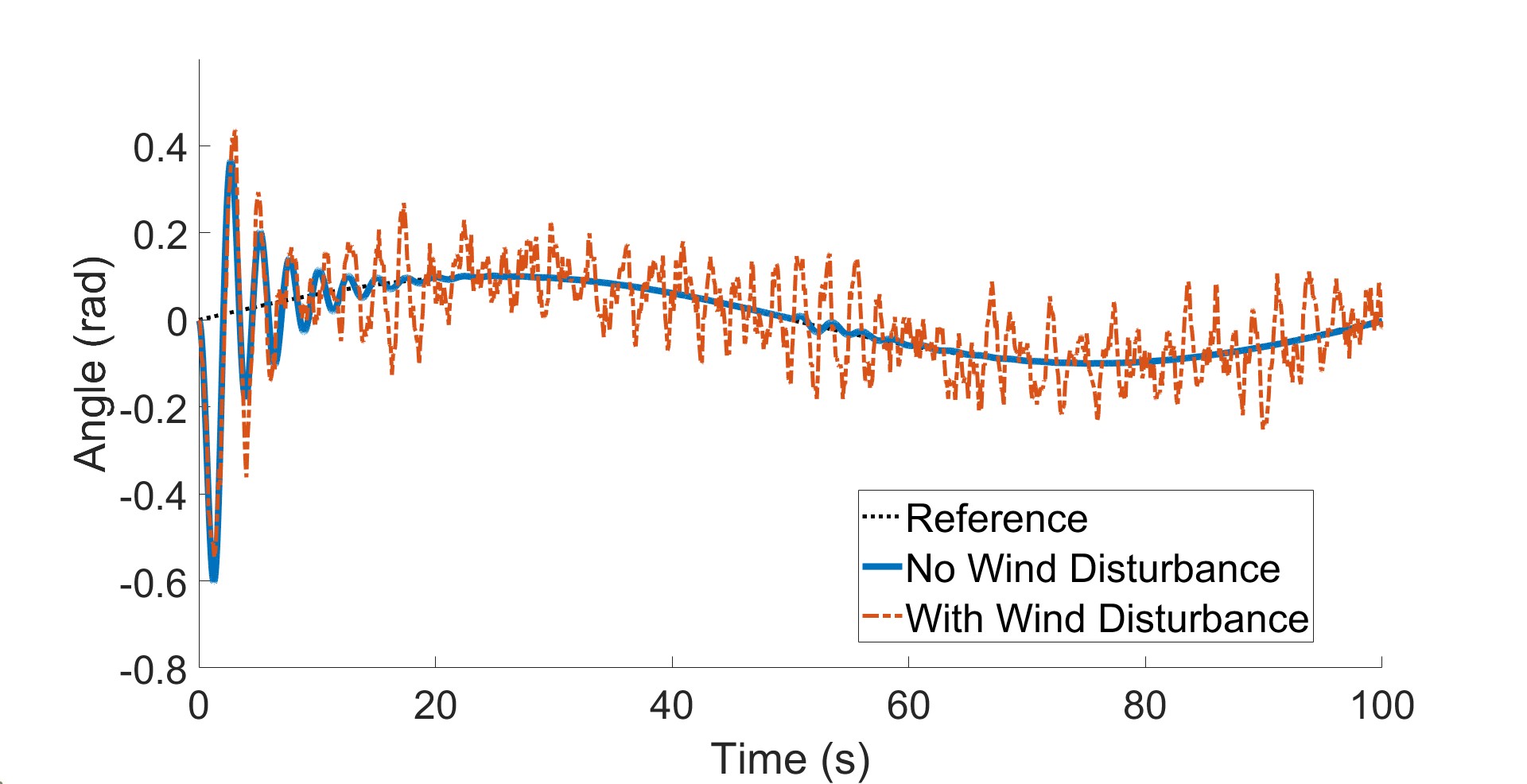}}
\caption{Pitch trajectory tracking response using PID controller}
\label{Pitch(pid)}
\end{figure}

Fig. \ref{Pitch(pid)} shows the pitch response for the PID controller. The controller without the disturbances was able to track the reference with minimal oscillations. With the disturbances, however, the PID controller tracked the reference with a lot of oscillations and didn't stabilize.

\begin{figure}[H]
\centerline{\includegraphics[width=0.5\textwidth]{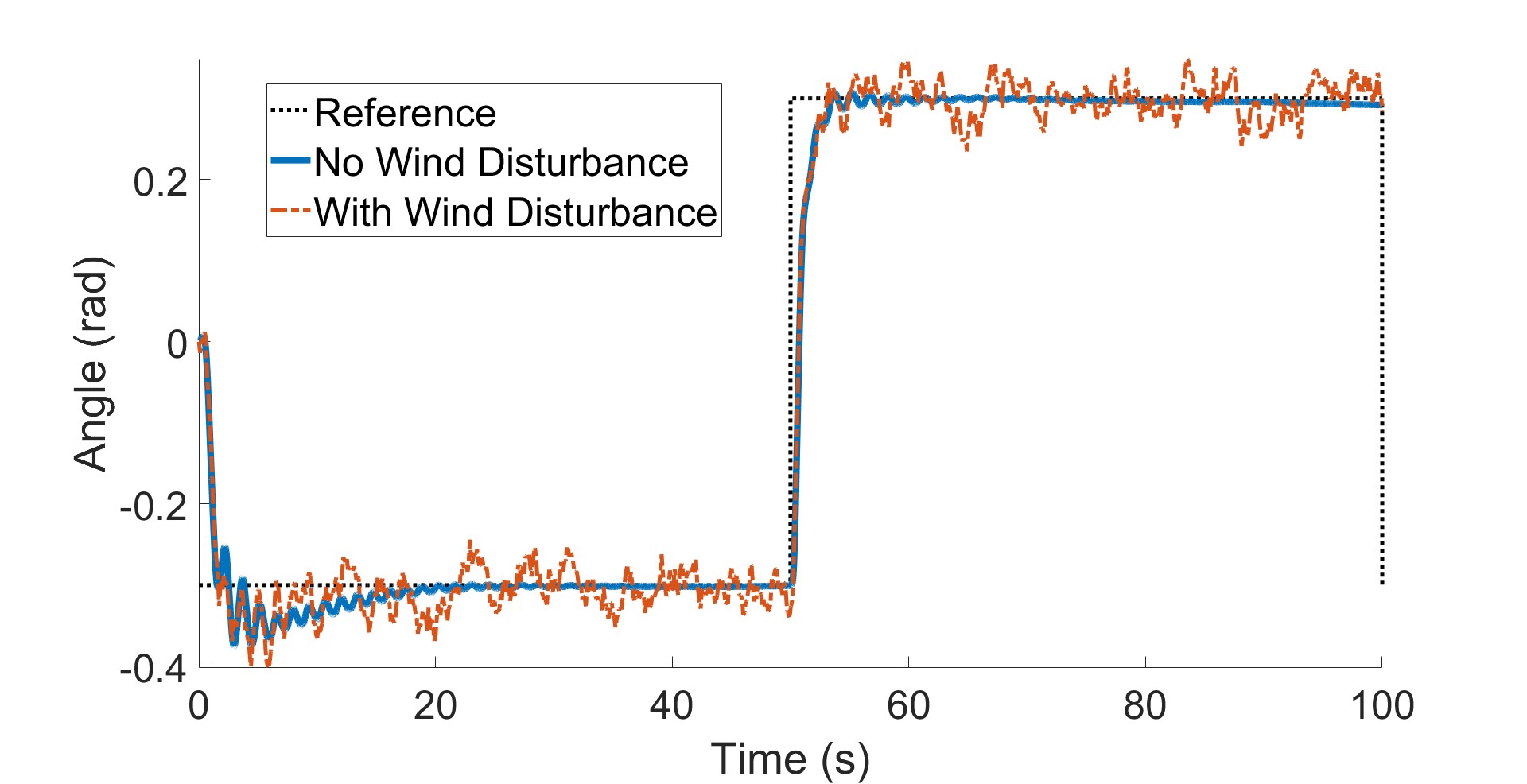}}
\caption{Azimuth trajectory tracking response using RL agent}
\label{Azimuth(RL)}
\end{figure}

Fig. \ref{Azimuth(RL)} shows the azimuth response when tracking the same trajectory but using RL. Without disturbances, the PID controller response settled faster than the RL but with a higher overshoot. However, when the disturbances were added, the RL agent was able to adapt better as the oscillations were lower than that of the PID controller.

\begin{figure}[H]
\centerline{\includegraphics[width=0.5\textwidth]{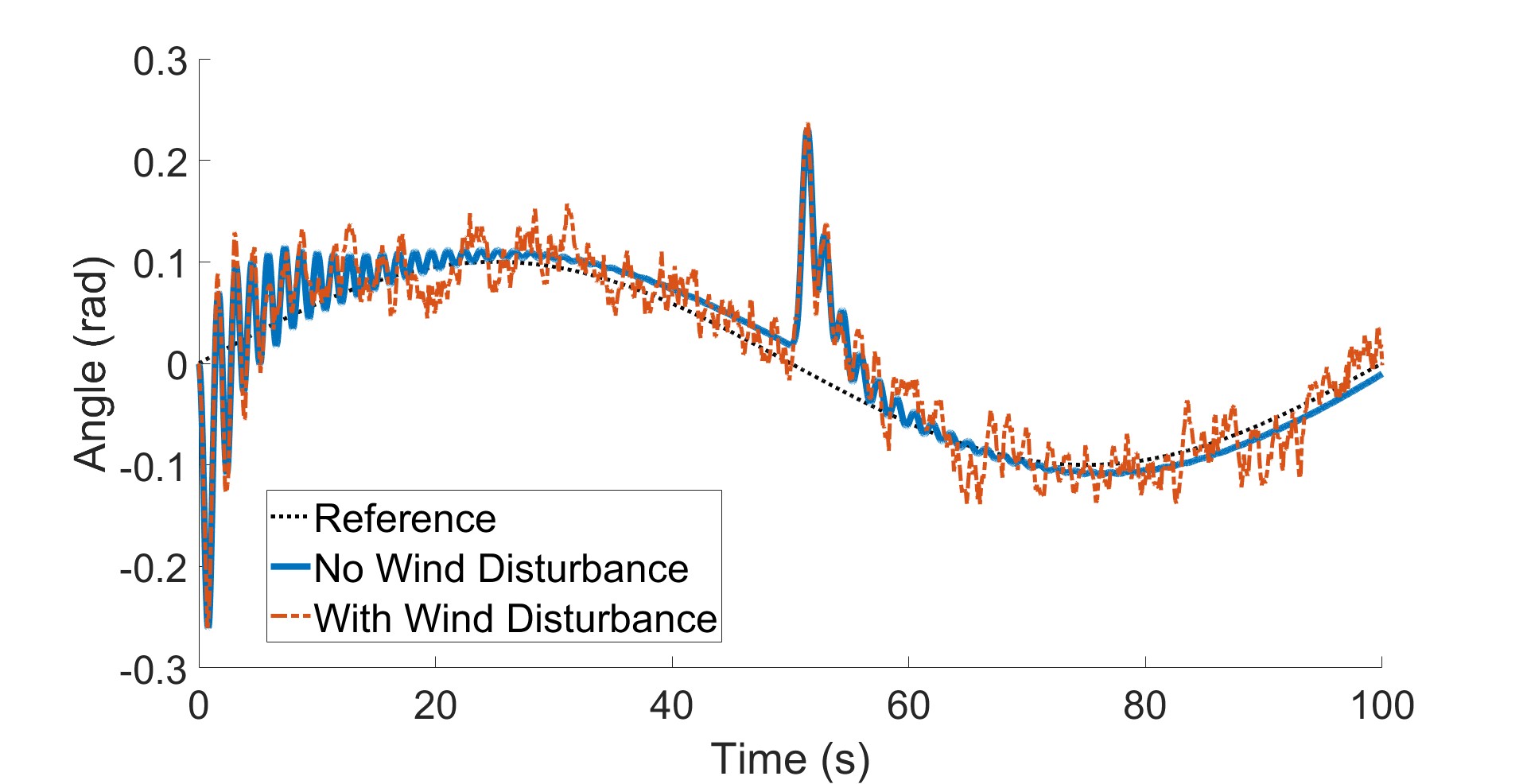}}
\caption{Pitch trajectory tracking response using RL agent}
\label{Pitch(RL)}
\end{figure}

Fig. \ref{Pitch(RL)} shows the pitch response using RL. Although the PID controller response is smoother and with lower steady-state error, with the disturbances added, the oscillations in the RL were lower in magnitude than that of the PID controller.

All things considered, the RL agent was more robust than the PID controller for this non-linear system as the PID controller is a linear controller with fixed gains and may not be able to adapt to changes in the dynamics or the parameters of the system due to external disturbances or other factors and the gains must be re-tuned. However, the RL agent can adapt its control strategy to show robustness in handling changes to the system. This is because the RL agent only evaluates the output states and not the dynamics of the system itself according to the optimal policy it learns through interactions with the environment.

\subsection{Experimental Validation}

\begin{figure}[H]
\centerline{\includegraphics[width=0.5\textwidth]{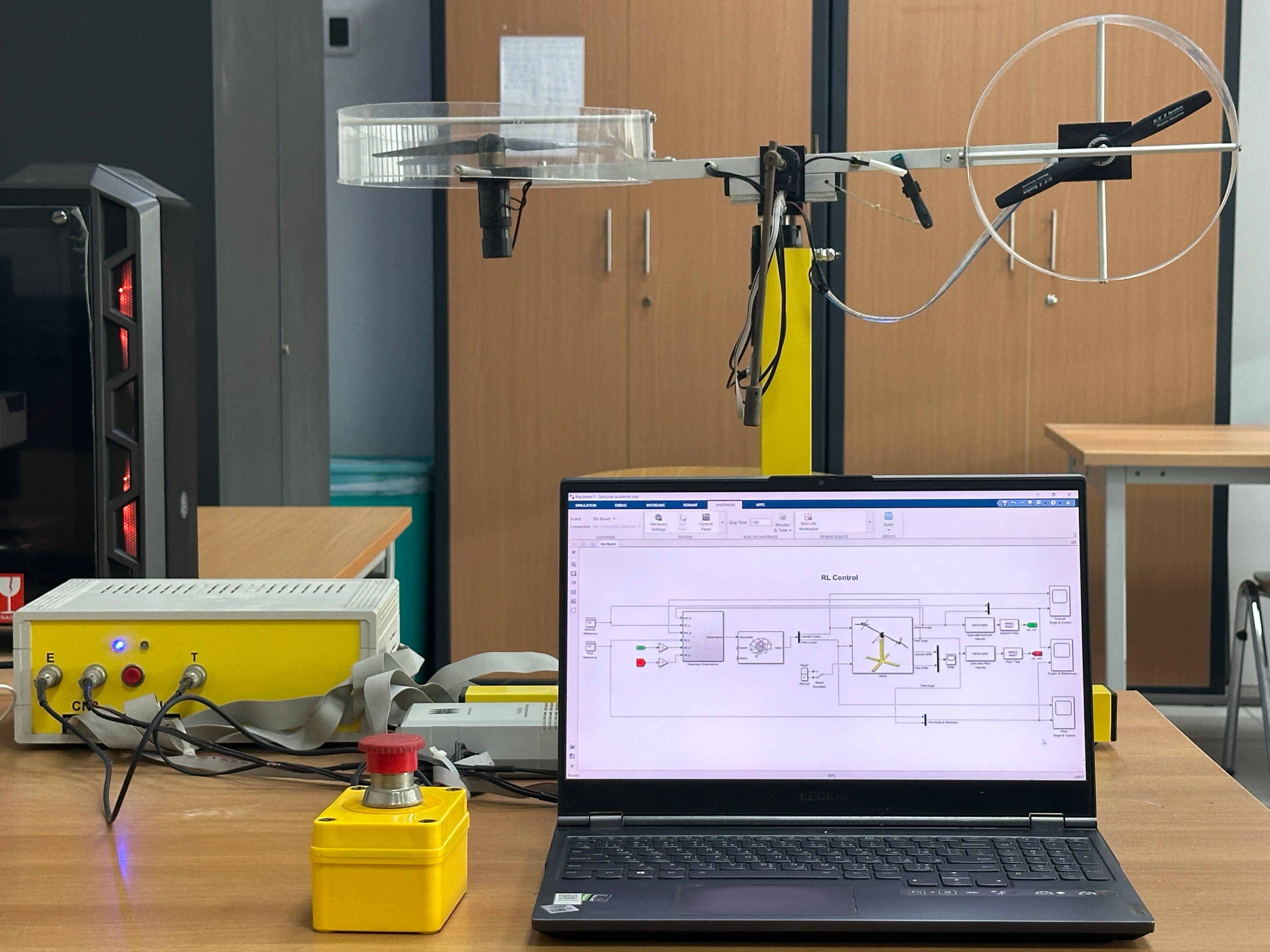}}
\caption{TRAS laboratory setup used in experiment}
\label{TRAS}
\end{figure}

After achieving good results when simulating different scenarios, experiments were then performed on the physical system to test the designed controller in real-time. The exact setup used is shown in Fig. \ref{TRAS}. A laboratory TRAS \cite{inteco} is connected to a measurement \& control FPGA device that sends physical data to the laptop connected to MATLAB/Simulink. The RL algorithm was then run on MATLAB/Simulink and the control signals were sent to the TRAS through the FPGA device.

\begin{figure}[H]
\centerline{\includegraphics[width=0.5\textwidth]{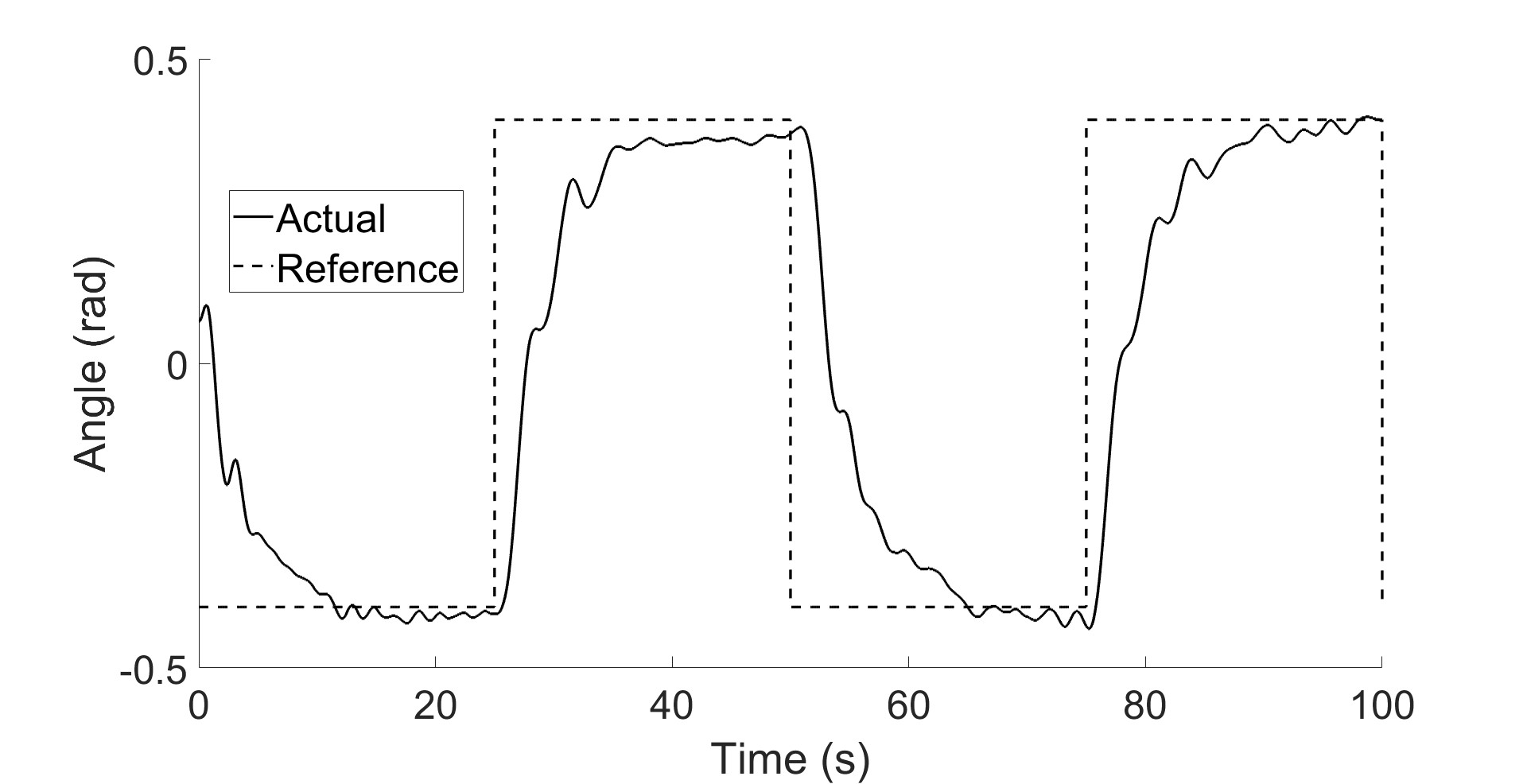}}
\caption{Azimuth trajectory tracking experimental results}
\label{Azimuth(hardware)}
\end{figure}

\begin{figure}[H]
\centerline{\includegraphics[width=0.5\textwidth]{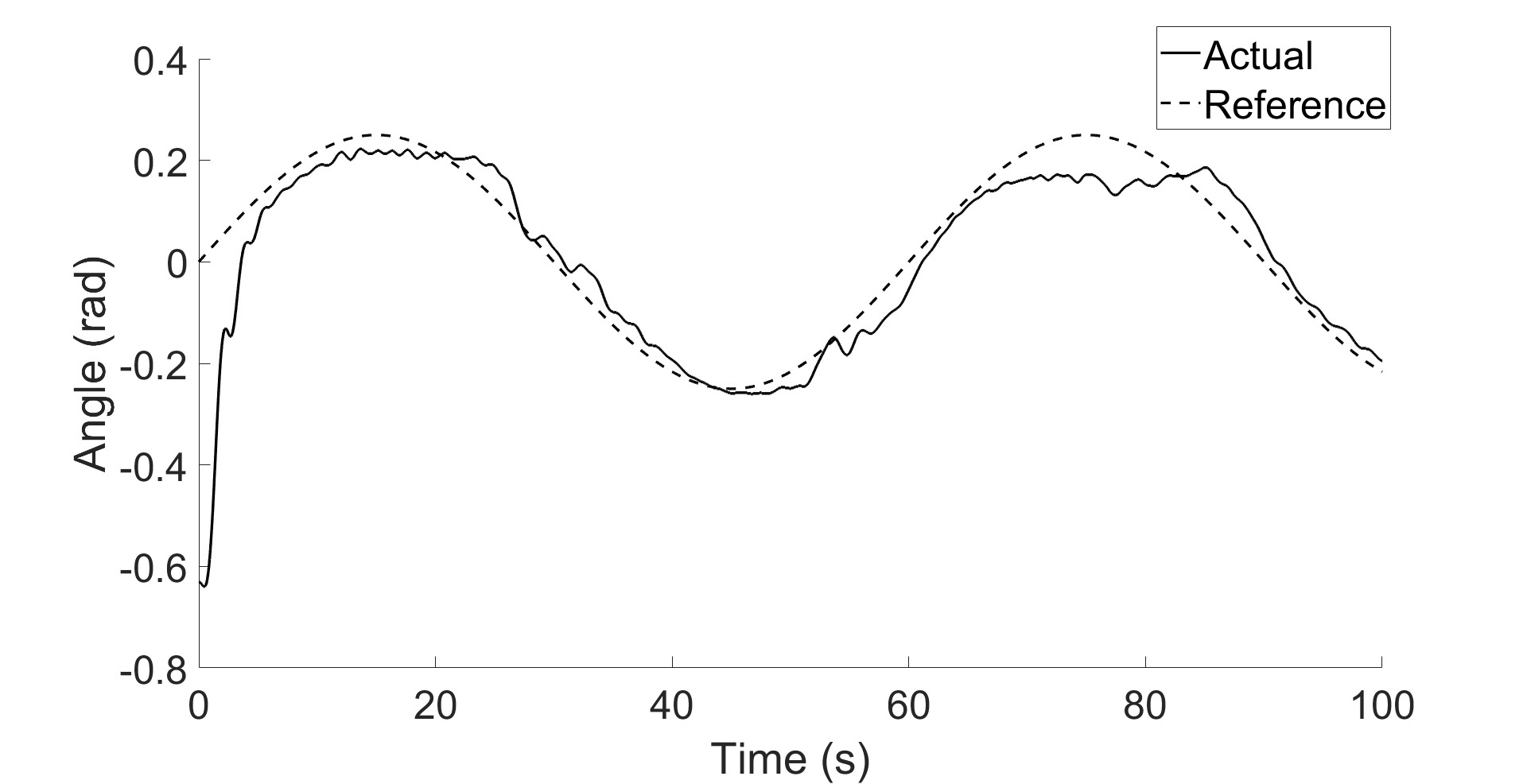}}
\caption{Pitch trajectory tracking experimental results}
\label{Pitch(hardware)}
\end{figure}

Fig. \ref{Azimuth(hardware)} and Fig. \ref{Pitch(hardware)} show that the real-time responses show satisfactory agreements with the simulation results, as they tracked the references with minimal error. There is an offset error compared to the simulation results because the agent was trained in an idealized simulated environment, which may not have accurately captured the dynamics of the actual system. When deployed on the real system, these differences became noticeable. To address this offset error, the agent could be further trained directly on the real system instead of relying solely on the simulated environment.

\section{Conclusion}
In this paper, a novel approach to controlling the TRAS
using deep reinforcement learning is presented. The RL agent
was trained using the TD3 algorithm to track different trajectories
and stabilize at any given position by mapping the states of TRAS and the tracking error into control signals to track the reference position.
The simulation results showed that the proposed controller
was able to effectively stabilize at any position and track different trajectories. Wind disturbances were added and the performance of the agent was compared to that of the PID
controller. Finally, experimental results were obtained by using
the agent to control a laboratory TRAS setup. The simulation
results prove the effectiveness and robustness of the RL agent
and its successful control of the TRAS in different scenarios.

\bibliographystyle{ieeetr}
\bibliography{references}

@inproceedings{mokhtar2023autonomous,
  title={Autonomous Navigation and Control of a Quadrotor Using Deep Reinforcement Learning},
  author={Mokhtar, Mohamed and El-Badawy, Ayman},
  booktitle={2023 International Conference on Unmanned Aircraft Systems (ICUAS)},
  pages={1045--1052},
  year={2023},
  organization={IEEE}
}

@inproceedings{hassan2020robust,
  title={Robust H-infinity Control for a Bi-rotor System},
  author={Hassan, Ragi and Hossam, Abdallah and El-Badawy, Ayman},
  booktitle={AIAA Scitech 2020 Forum},
  pages={1834},
  year={2020}
}

@article{juang2011hybrid,
  title={A hybrid intelligent controller for a twin rotor MIMO system and its hardware implementation},
  author={Juang, Jih-Gau and Liu, Wen-Kai and Lin, Ren-Wei},
  journal={ISA transactions},
  volume={50},
  number={4},
  pages={609--619},
  year={2011},
  publisher={Elsevier}
}

@inproceedings{ahmed2009robust,
  title={Robust decoupling control design for twin rotor system using Hadamard weights},
  author={Ahmed, Qadeer and Bhatti, Aamer Iqbal and Iqbal, Sohail},
  booktitle={2009 IEEE Control Applications,(CCA) \& Intelligent Control,(ISIC)},
  pages={1009--1014},
  year={2009},
  organization={IEEE}
}

@article{rashad2017sliding,
  title={Sliding mode disturbance observer-based control of a twin rotor MIMO system},
  author={Rashad, Ramy and El-Badawy, Ayman and Aboudonia, Ahmed},
  journal={ISA transactions},
  volume={69},
  pages={166--174},
  year={2017},
  publisher={Elsevier}
}

@inproceedings{shehab2021low,
  title={Low-level control of a quadrotor using twin delayed deep deterministic policy gradient (td3)},
  author={Shehab, Mazen and Zaghloul, Ahmed and El-Badawy, Ayman},
  booktitle={2021 18th International Conference on Electrical Engineering, Computing Science and Automatic Control (CCE)},
  pages={1--6},
  year={2021},
  organization={IEEE}
}

@article{hwangbo2017control,
  title={Control of a quadrotor with reinforcement learning},
  author={Hwangbo, Jemin and Sa, Inkyu and Siegwart, Roland and Hutter, Marco},
  journal={IEEE Robotics and Automation Letters},
  volume={2},
  number={4},
  pages={2096--2103},
  year={2017},
  publisher={IEEE}
}

@misc{inteco,
  title = {Two Rotor Aerodynamical System, User's Manual},
  howpublished = {\url{http://www.inteco.com.pl/}},
  year = {2006}
}

@inproceedings{fujimoto2018addressing,
  title={Addressing function approximation error in actor-critic methods},
  author={Fujimoto, Scott and Hoof, Herke and Meger, David},
  booktitle={International conference on machine learning},
  pages={1587--1596},
  year={2018},
  organization={PMLR}
}

@inproceedings{ulasyar2015robust,
  title={Robust \& optimal model predictive controller design for twin rotor MIMO system},
  author={Ulasyar, Abasin and Zad, Haris Sheh},
  booktitle={2015 9th International Conference on Electrical and Electronics Engineering (ELECO)},
  pages={854--858},
  year={2015},
  organization={IEEE}
}

@misc{moorhouse1982background,
  title={Background Information and User Guide for MIL-F-8785C, Military Specification: Flying Qualities of Piloted Airplanes},
  author={Moorhouse, David J and Woodcock, Robert J},
  year={1982},
  publisher={Air Force Flight Dynamics Laboratory}
}
\end{document}